\documentclass[11pt,a4paper]{article}
\usepackage[hyperref]{eacl2021}
\usepackage{times}
\usepackage{latexsym}
\usepackage{booktabs}
\usepackage{tabularx}
\usepackage{multicol}
\usepackage{multirow}
\usepackage{graphicx}
\usepackage{caption}
\usepackage{subcaption}
\usepackage{fdsymbol}

\usepackage[compact]{titlesec}

\usepackage{microtype}

\aclfinalcopy 
\def\aclpaperid{2759} 

\title{Pre-training is a Hot Topic: Contextualized Document Embeddings Improve Topic Coherence}

\author{Federico Bianchi \\
    Bocconi University \\ 
    Via Sarfatti 25, 20136 \\ Milan, Italy \\  f.bianchi@unibocconi.it \\\And
  Silvia Terragni \\
    University of Milan-Bicocca \\
    Viale Sarca 336, 20126 \\
    Milan, Italy  \\
       s.terragni4@campus.unimib.it\\ \And
  Dirk Hovy \\
    Bocconi University \\ 
    Via Sarfatti 25, 20136 \\ Milan, Italy \\
    dirk.hovy@unibocconi.it}

\date{}

\begin{document}
\maketitle
\begin{abstract}
Topic models extract groups of words from documents, whose interpretation as a topic hopefully allows for a better understanding of the data. However, the resulting word groups are often not \textit{coherent}, making them harder to interpret. 
Recently, neural topic models have shown improvements in overall coherence. Concurrently, contextual embeddings have advanced the state of the art of neural models in general.
In this paper, we combine contextualized representations with neural topic models. We find that our approach produces more meaningful and coherent topics than traditional bag-of-words topic models and recent neural models. Our results indicate that future improvements in language models will translate into better topic models.
\end{abstract}


\section{Introduction}
One of the crucial issues with topic models is the quality of the topics they discover. \textit{Coherent} topics are easier to interpret and are considered more meaningful. E.g., a topic represented by the words ``apple, pear, lemon, banana, kiwi'' would be considered a meaningful topic on \textit{FRUIT} and is more coherent than one defined by ``apple, knife, lemon, banana, spoon.'' Coherence can be measured in numerous ways, from human evaluation via intrusion tests~\cite{Chang09topicintrusion} to approximated scores~\cite{Lau14npmi,Roder15coherence}.

However, most topic models still use Bag-of-Words (BoW) document representations as input. These representations, though, disregard the syntactic and semantic relationships among the words in a document, the two main linguistic avenues to coherent text. I.e., BoW models represent the input in an inherently incoherent manner.

Meanwhile, pre-trained language models are becoming ubiquitous in Natural Language Processing (NLP), precisely for their ability to capture and maintain sentential coherence. 
Bidirectional Encoder Representations from Transformers (BERT)~\cite{devlin2018bert}, the most prominent architecture in this category, allows us to extract pre-trained word and sentence representations. Their use as input has advanced state-of-the-art performance across many tasks. Consequently, BERT representations are used in a diverse set of NLP applications~\cite{rogers2020primer,nozza2020mask}.

Various extensions of topic models incorporate several types of information~\cite{Xun17correlatedtm,zhao2017metalda,terragni2020constrained}, use word relationships derived from external knowledge bases~\cite{Chen13externalknowledgelda,Yang15clda,terragni2020matters}, or pre-trained word embeddings~\cite{Das15gaussianlda,dieng2019diversity,Nguyen15lflda,zhao2017metalda}. Even for neural topic models, there exists work on incorporating external knowledge, e.g., via word embeddings~\cite{Gupta19idocnade,gupta2020continualtm,dieng2019diversity}. 

In this paper, we show that adding contextual information to neural topic models provides a \textbf{significant} increase in topic coherence. This effect is even more remarkable given that we cannot embed long documents due to the sentence length limit in recent pre-trained language models architectures. 

Concretely, we extend Neural ProdLDA (Product-of-Experts LDA)~\cite{srivastava2017avitm}, a state-of-the-art topic model that implements black-box variational inference~\cite{Ranganath14blackbox}, to include contextualized representations.
Our approach leads to consistent and significant improvements in two standard metrics on topic coherence and produces competitive topic diversity results. 

\paragraph{Contributions} We propose a straightforward and easily implementable method that allows neural topic models to create coherent topics. 
We show that the use of contextualized document embeddings in neural topic models produces significantly more coherent topics.
Our results suggest that topic models benefit from latent contextual information, which is missing in BoW representations. The resulting model addresses one of the most central issues in topic modeling. We release our implementation as a Python library, available at the following link: \url{https://github.com/MilaNLProc/contextualized-topic-models}.

\section{Neural Topic Models with Language Model Pre-training}
We introduce a Combined Topic Model (CombinedTM) to investigate the incorporation of contextualized representations in topic models. Our model is built around two main components: (i)  the neural topic model ProdLDA~\cite{srivastava2017avitm} and (ii) the SBERT embedded representations~\cite{reimers2019sentence}. Let us notice that our method is indeed agnostic about the choice of the topic model and the pre-trained representations, as long as the topic model extends an autoencoder and the pre-trained representations embed the documents.

ProdLDA is a neural topic modeling approach based on the Variational AutoEncoder (VAE). The neural variational framework trains a neural inference network to directly map the BoW document representation into a continuous latent representation. Then, a decoder network reconstructs the BoW by generating its words from the latent document representation\footnote{For more details see ~\cite{srivastava2017avitm}.}.  The framework explicitly approximates the Dirichlet prior using Gaussian distributions, instead of using a Gaussian prior like Neural Variational Document Models~\cite{Miao2015NVDM}.
Moreover, ProdLDA replaces the multinomial distribution over individual words in LDA with a product of experts~\cite{Hinton02prod}.

\begin{figure}[ht]
\centering
    \includegraphics[width=0.9\columnwidth]{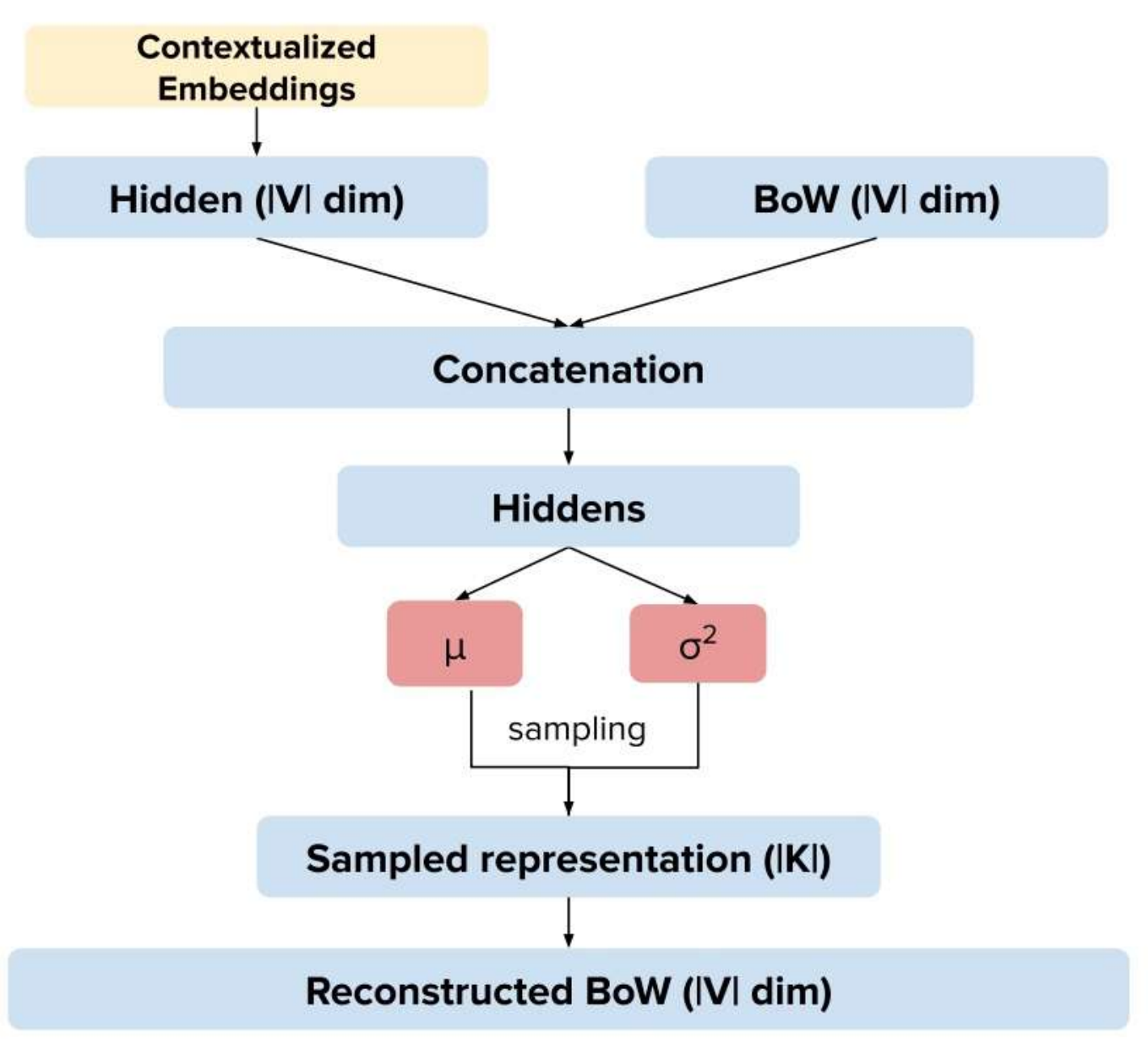}
    \caption{High-level sketch of CombinedTM. Refer to~\cite{srivastava2017avitm} for more details on the architecture we extend.}
    \label{fig:lm_topic}
\end{figure}

We extend this model with contextualized document embeddings from SBERT~\cite{reimers2019sentence},\footnote{\url{https://github.com/UKPLab/sentence-transformers}} a recent extension of BERT that allows the quick generation of sentence embeddings. This approach has one limitation. If a document is longer than SBERT's sentence-length limit, the rest of the document will be lost. 
The document representations are projected through a hidden layer with the same dimensionality as the vocabulary size, concatenated with the BoW representation.  Figure~\ref{fig:lm_topic} briefly sketches the architecture of our model. The hidden layer size could be tuned, but an extensive evaluation of different architectures is out of the scope of this paper.

{}

\begin{table}[ht]
    \centering
    \begin{tabular}{l|rr} \toprule
         Dataset & Docs & Vocabulary \\ \midrule
         20Newsgroups & 18,173 & 2,000 \\ 
         Wiki20K & 20,000 & 2,000 \\
         StackOverflow & 16,408 & 2,303 \\
         Tweets2011 & 2,471 & 5,098 \\
         Google News & 11,108 & 8,110 
\\ \bottomrule 
    \end{tabular}
    \caption{Statistics of the datasets used.}
    \label{tab:dataset:statistics}
\end{table}{}

\section{Experimental Setting}
We provide detailed explanations of the experiments (e.g., runtimes) in the supplementary materials. To reach full replicability, we use open-source implementations of the algorithms. 

\begin{table}[ht!]
    \centering
 
    \begin{tabular}{l|rrr} \toprule
        \textbf{Model} & \textbf{Avg $\tau$} & \textbf{Avg $\alpha$}& \textbf{Avg $\rho$} \\ 
        \midrule
    \multicolumn{4}{c}{Results for the Wiki20K Dataset:
    \label{tab:wiki}}\\
        \midrule
        \textbf{Ours} & \textbf{0.1823}   &  0.1980 &  \textbf{0.9950}  \\ 
        PLDA & 0.1397  & 0.1799 & 0.9901  \\ 
        MLDA & 0.1443 & \textbf{0.2110} &  0.9843 \\
        NVDM  & -0.2938    &  0.0797  & 0.9604  \\
        ETM &  0.0740  &	0.1948  & 0.8632  \\
        LDA & -0.0481 &0.1333  & 0.9931  \\
        \midrule

    \multicolumn{4}{c}{Results for the StackOverflow Dataset:
    \label{tab:stack}}\\ \midrule
         \textbf{Ours} &\textbf{0.0280} & 0.1563  & 0.9805   \\ 
        PLDA & -0.0394  & 0.1370 & \textbf{0.9914}\\ 
        MLDA & 0.0136 & 0.1450 & 0.9822 \\
        NVDM &  -0.4836    & 0.0985 &    0.8903 \\
        ETM & -0.4132 & \textbf{0.1598} &	0.4788	\\
        LDA & -0.3207     & 0.1063  & 0.8947\\ 
        \midrule
        \multicolumn{4}{c}{Results for the GoogleNews Dataset:
    \label{tab:gnews}}\\
    \midrule
     \textbf{Ours} & \textbf{0.1207} &  \textbf{0.1325} & \textbf{0.9965}  \\ 
        PLDA & 0.0110 & 0.1218   & 0.9902\\ 
        MLDA & 0.0849 & 0.1219 & 0.9959 \\
        NVDM  &  -0.3767& 0.1067 & 0.9648  \\
        ETM & -0.2770 &	0.1175 & 0.4700\\
        LDA & -0.3250 & 0.0969 & 0.9774  \\ 
        \midrule
    \multicolumn{4}{c}{Results for the Tweets2011 Dataset:  \label{tab:tweets}}\\ \midrule
    \textbf{Ours} & \textbf{0.1008} & \textbf{0.1493} & 0.9901 \\ 
        PLDA & 0.0612  & 0.1327 & 0.9847 \\ 
        MLDA & 0.0122 & 0.1272 & \textbf{0.9956} \\
        NVDM  & -0.5105  & 0.0797 & 0.9751 \\
        ETM & -0.3613 &	0.1166 &	0.4335 \\
        LDA & -0.3227 & 0.1025 & 0.8169  \\ \midrule

        \multicolumn{4}{c}{Results for the 20NewsGroups Dataset:
    \label{tab:20}}\\
        \midrule
        \textbf{Ours} & 0.1025 & 0.1715 &  0.9917    \\ 
        PLDA & 0.0632 & 0.1554 & \textbf{0.9931} \\ 
        MLDA & \textbf{0.1300} & 0.2210 & 0.9808 \\
        NVDM & -0.1720 & 0.0839& 0.9805  \\
        ETM & 0.0766   & \textbf{0.2539} &    0.8642 \\
        LDA & 0.0173  & 0.1627 & 0.9897 \\ 
        \bottomrule
    \end{tabular}
    \caption{Averaged results over 5 numbers of topics. Best results are marked in bold.} \label{tab:avg_res}
\end{table}

\subsection{Datasets}
We evaluate the models on five datasets: 20NewsGroups\footnote{\url{http://qwone.com/~jason/20Newsgroups/}}, Wiki20K (a collection of 20,000 English Wikipedia abstracts from~\citet{bianchi2020cross}), Tweets2011\footnote{\url{https://trec.nist.gov/data/tweets/}}, Google News~\cite{qiang2018STTP}, and the StackOverflow dataset~\cite{qiang2018STTP}. The latter three are already pre-processed. We use a similar pipeline for 20NewsGroups and Wiki20K: removing digits, punctuation, stopwords, and infrequent words. We derive SBERT document representations from unpreprocessed text for Wiki20k and 20NewsGroups. For the others, we use the pre-processed text;\footnote{This can be sub-optimal, but many datasets in the literature are already pre-processed.}  See Table~\ref{tab:dataset:statistics} for dataset statistics. The sentence encoding model used is the pre-trained RoBERTa model fine-tuned on SNLI~\cite{bowman-etal-2015-large}, MNLI~\cite{williams-etal-2018-broad}, and the STSb~\cite{cer-etal-2017-semeval} dataset.\footnote{stsb-roberta-large}

\subsection{Metrics} We evaluate each model on three different metrics: two for topic coherence (normalized pointwise mutual information and a word-embedding based measure) and one metric to quantify the diversity of the topic solutions.

\paragraph{Normalized Pointwise Mutual Information ($\tau$)}~\cite{Lau14npmi} 
measures how related the top-10 words of a topic are to each other, considering the words' empirical frequency in the original corpus. $\tau$ is a symbolic metric and relies on co-occurrence. As \newcite{ding2018coherence} pointed out, though, topic coherence computed on the original data is inherently limited. Coherence computed on an external corpus, on the other hand, correlates much more to human judgment, but it may be expensive to estimate. 

\paragraph{External word embeddings topic coherence ($\alpha$)} 
provides an additional measure of how similar the words in a topic are. We follow \newcite{ding2018coherence} and first compute the average pairwise cosine similarity of the word embeddings of the top-10 words in a topic, using~\newcite{mikolov2013distributed} embeddings. Then, we compute the overall average of those values for all the topics. We can consider this measure as an external topic coherence, but it is more efficient to compute than Normalized Pointwise Mutual Information on an external corpus. 

\paragraph{Inversed Rank-Biased Overlap ($\rho$)} evaluates how diverse the topics generated by a single model are.
We define $\rho$ as the reciprocal of the standard RBO~\cite{webber2010similarity,terragni2021similarity}. RBO compares the 10-top words of two topics. It allows disjointedness between the lists of topics (i.e., two topics can have different words in them) and uses weighted ranking. I.e., two lists that share some of the same words, albeit at different rankings, are penalized less than two lists that share the same words at the highest ranks. 
$\rho$ is 0 for identical topics and 1 for completely different topics. 

\subsection{Models} 
Our main objective is to show that contextual information increases coherence. To show this, we compare our approach to ProdLDA~\cite[the model we extend]{srivastava2017avitm}\footnote{We use the implementation of \newcite{Carrow2018avitm}.}, and the following models:  (ii) Neural Variational Document Model (NVDM)~\cite{Miao2015NVDM}; (iii) the very recent ETM~\cite{dieng2019diversity}, MetaLDA (MLDA)~\cite{zhao2017metalda} and (iv) LDA~\cite{Blei03lda}.

\subsection{Configurations}

To maximize comparability, we train all models with similar hyper-parameter configurations. The inference network for both our method and ProdLDA consists of one hidden layer and 100-dimension of softplus units, which converts the input into embeddings. This final representation is again passed through a hidden layer before the variational inference process. We follow \cite{srivastava2017avitm} for the choice of the parameters. The priors over the topic and document distributions are learnable parameters. For LDA, the Dirichlet priors are estimated via Expectation-Maximization. See the Supplementary Materials for additional details on the configurations.


\section{Results}
We divide our results into two parts: we first describe the results for our quantitative evaluation, and we then explore the effect on the performance when we use two different contextualized representations. 

\subsection{Quantitative Evaluation}
We compute all the metrics for 25, 50, 75, 100, and 150 topics. We average results for each metric over 30 runs of each model (see Table~\ref{tab:avg_res}).

As a general remark, our model provides the most coherent topics across all corpora and topic settings, even maintaining a competitive diversity of the topics. This result suggests that the incorporation of contextualized representations can improve a topic model's performance. 

LDA and NVDM obtain low coherence. This result has also also been confirmed by  \newcite{srivastava2017avitm}. 
ETM shows good external coherence ($\alpha$), especially in 20NewsGroups and StackOverflow. However, it fails at obtaining a good $\tau$ coherence for short texts. Moreover, $\rho$ shows that the topics are very similar to one another. A manual inspection of the topics confirmed this problem. MetaLDA is the most competitive of the models we used for comparison. This may be due to the incorporation of pre-trained word embeddings into MetaLDA. Our model provides very competitive results, and the second strongest model appears to be MetaLDA. For this reason, we provide a detailed comparison of $\tau$ in Table~\ref{tab:mldavsours}, where we show the average coherence for each number of topics; we show that on 4 datasets over 5 our model provides the best results, but still keeping a very competitive score on 20NG, where MetaLDA is best.

Readers can see examples of the top words for each model in the Supplementary Materials. These descriptors illustrate the increased coherence of topics obtained with SBERT embeddings.

\begin{table}[]
    \centering
    \small
    \begin{tabular}{l|lllll} \toprule
        \textbf{Wiki20K}  & \multicolumn{1}{c}{25} & \multicolumn{1}{c}{50}  & \multicolumn{1}{c}{75}  & \multicolumn{1}{c}{100}  & \multicolumn{1}{c}{150}   \\ \midrule
        Ours &  \textbf{0.17}$^\clubsuit$ & \textbf{0.19}$^\clubsuit$  & \textbf{0.18}$^\clubsuit$  & \textbf{0.19}$^\clubsuit$  & \textbf{0.17}$^\clubsuit$  \\
        MLDA & 0.15 & 0.15 & 0.14 & 0.14 & 0.13 \\ \midrule
        \textbf{SO}    \\ \midrule
        Ours & \textbf{0.05} & \textbf{0.03}$^\clubsuit$ & \textbf{0.02}$^\clubsuit$  & \textbf{0.02}$^\clubsuit$  & \textbf{0.02}$^\clubsuit$  \\
        MLDA & \textbf{0.05}$^\clubsuit$ & 0.02 & 0.00 & -0.02 & 0.00 \\ \midrule
        \textbf{GNEWS}    \\ \midrule
        Ours & \textbf{-0.03}$^\clubsuit$  & \textbf{0.10}$^\clubsuit$  & \textbf{0.15}$^\clubsuit$  & \textbf{0.18}$^\clubsuit$  & \textbf{0.19}$^\clubsuit$ \\
        MLDA & -0.06 & 0.07 & 0.13 & 0.16 & 0.14\\ \midrule
        \textbf{Tweets}  &   \\ \midrule
        Ours & \textbf{0.05}$^\clubsuit$ & \textbf{0.10}$^\clubsuit$  & \textbf{0.11}$^\clubsuit$  & \textbf{0.12}$^\clubsuit$  & \textbf{0.12}$^\clubsuit$   \\
        MLDA & 0.00 & 0.05& 0.06& 0.04 & -0.07  \\ \midrule
                \textbf{20NG}   \\ \midrule
        Ours & 0.12 & 0.11 &0.10 &0.09 & 0.09   \\
        MLDA & \textbf{0.13}$^\clubsuit$  & \textbf{0.13}$^\clubsuit$  &\textbf{0.13}$^\clubsuit$  &\textbf{0.13}$^\clubsuit$  &\textbf{0.12}$^\clubsuit$ \\ \bottomrule
    \end{tabular}
    \caption{Comparison of $\tau$ between CombinedTM (ours) and MetaLDA over various choices of topics. Each result averaged over 30 runs. $^\clubsuit$ indicates statistical significance of the results (t-test, p-value $<0.05$).}
    \label{tab:mldavsours}
\end{table}


\subsection{Using Different Contextualized Representations}
Contextualized representations can be generated from different models and some representations might be better than others. Indeed, one question left to answer is the impact of the specific contextualized model on the topic modeling task.
To answer to this question we rerun all the experiments with CombinedTM but we used different contextualized sentence embedding methods as input to the model.

We compare the performance of CombinedTM using two different models for embedding the contextualized representations found in the SBERT repository:\footnote{\url{https://github.com/UKPLab/sentence-transformers}} \textit{stsb-roberta-large} (Ours-R), as employed in the previous experimental setting, and using \textit{bert-base-nli-means} (Ours-B). The latter is derived from a BERT model fine-tuned on NLI data.
Table~\ref{tab:ablation} shows the coherence of the two approaches on all the datasets (we averaged all results). In these experiments, RoBERTa fine-tuned on the STSb dataset has a strong impact on the increase of the coherence. This result suggests that including novel and better contextualized embeddings can further improve a topic model's performance.

\begin{table}[]
    \centering
    \small
    \begin{tabular}{cccccc} \toprule
        & \textbf{Wiki20K} & \textbf{SO} & \textbf{GN} & \textbf{Tweets} & \textbf{20NG}  \\ \midrule
        Ours-R & \textbf{0.18} & \textbf{0.03}  & \textbf{0.12} & \textbf{0.10} & \textbf{0.10}  \\
        Ours-B & \textbf{0.18} & 0.02 & 0.08 & 0.06 & 0.07 \\ \bottomrule
    \end{tabular}
    \caption{$\tau$ performance of CombinedTM using different contextualized encoders.}
    \label{tab:ablation}
\end{table}

\section{Related Work}
In recent years, neural topic models have gained increasing success and interest~\cite{zhao2021survey,terragni2021octis}, due to their flexibility and scalability.  
Several topic models use neural networks \cite{Larochelle12autoregressive,Salakhutdinov09replicatedsoftmax,gupta2020continualtm} or neural variational inference \cite{Miao2015NVDM, Mnih14neuralvarinference, srivastava2017avitm,Miao17discreteneuralvariationalinf,ding2018coherence}. \newcite{Miao2015NVDM} propose NVDM, an unsupervised generative model based on VAEs, assuming a Gaussian distribution over topics. Building upon NVDM, \citet{dieng2019diversity} represent words and topics in the same embedding space.
\newcite{srivastava2017avitm} propose a neural variational framework that explicitly approximates the Dirichlet prior using a Gaussian distribution. 
Our approach builds on this work but includes a crucial component, i.e., the representations from a pre-trained transformer that can benefit from both general
language knowledge and corpus-dependent information. Similarly, \newcite{bianchi2020cross} replace the BOW document representation with pre-trained contextualized representations to tackle a problem of cross-lingual zero-shot topic modeling. This approach was extended by \newcite{mueller-dredze-2021-fine} that also considered fine-tuning the representations. A very recent approach \cite{hoyle2020distillation} which follows a similar direction uses knowledge distillation \cite{hinton2015distillation} to combine neural topic models and pre-trained transformers. 

\section{Conclusions}
We propose a straightforward and simple method to incorporate contextualized embeddings into topic models. The proposed model significantly improves the quality of the discovered topics. Our results show that context information is a significant element to consider also for topic modeling.

\section*{Ethical Statement}
In this research work, we used datasets from the recent literature, and we do not use or infer any sensible information. The risk of possible abuse of our models is low.

\section*{Acknowledgments}
We thank our colleague Debora Nozza and Wray Buntine for providing insightful comments on a previous version of this paper. Federico Bianchi and Dirk Hovy are members of the Bocconi Institute for Data Science and Analytics (BIDSA) and the Data and Marketing Insights (DMI) unit.


\bibliography{emnlp2020}
\bibliographystyle{acl_natbib}

\appendix

\section{Datasets}

We pre-processed 20NewsGroup and Wiki20K.  We removed punctuation, digits, and nltk's English stop-words. Following other researchers, we selected 2,000 as the maximum number of words for the BoW, and thus we kept only the 2,000 most frequent words in the documents. The other datasets come already pre-processed (reference links are in the paper) and thus we take them as is. 

\section{Models and Baselines}

\subsection{ProdLDA} We use the implementation made available by~\newcite{Carrow2018avitm} since it is the most recent and with the most updated packages (e.g., one of the latest versions of PyTorch). We run 100 epochs of the model. We use ADAM optimizer. The inference network is composed of a single hidden layer and 100-dimension of softplus units. The priors over the topic and document distributions are learnable parameters. Momentum is set to 0.99, the learning rate is set to 0.002, and we apply 20\% of drop-out to the hidden document representation. The batch size is equal to 200. More details related to the architecture can be found in the original work~\cite{srivastava2017avitm}.

\subsection{Combined TM} The model and the hyper-parameters are the same used for ProdLDA with the difference that we also use SBERT features in combination with the BoW: we take the SBERT English embeddings, apply a (learnable) function/dense layer $R^{1024} \rightarrow R^{|V|}$ and concatenate the representation to the BoW. We run 100 epochs of the model. We use ADAM optimizer.

\subsection{LDA} We use Gensim's\footnote{\url{https://radimrehurek.com/gensim/models/ldamodel.html}} implementation of this model. The hyper-parameters $\alpha$ and $\beta$, controlling the document-topic and word-topic distribution respectively, are estimated from the data during training.

\subsection{ETM}
We use the implementation available at \url{https://github.com/adjidieng/ETM} with default hyperparameters.

\subsection{Meta-LDA}
We use the authors' implementation available at \url{https://github.com/ethanhezhao/MetaLDA}. As suggested, we use the Glove embeddings to initialize the models. We used the 50-dimensional embeddings from \url{https://nlp.stanford.edu/projects/glove/}. The parameters $\alpha$ and $\beta$ have been set to 0.1 and 0.01 respectively.

\subsection{Neural Variational Document Model (NVDM)}
We use the implementation available at \url{https://github.com/ysmiao/nvdm} with default hyperparameters, but using two alternating epochs for encoder and decoder.

\section{Computing Infrastructure}
We ran the experiments on two common laptops, equipped with a GeForce GTX 1050: models can be easily run with basic infrastructure (having a GPU is better than just using CPU, but the experiment can also be replicated with CPU). Both laptops have 16GB of RAM. CUDA version for the experiments was 10.0.

\subsection{Runtime}

What influences the computational time the most is the number of words in the vocabulary. Table~\ref{tab:my_label} shows the runtime for one epoch of both our Combined TM (CTM) and ProdLDA (PDLDA) for 25 and 50 topics on Google News and 20Newsgroups datasets with the GeForce GTX 1050. ProdLDA is faster than our Combined TM. This is due to the added representation. However, we believe that these numbers are quite similar and make our model easy to use, even with common hardware.

\begin{table}[h]
\centering
    \small
\begin{tabular}{@{}c|cc|cc@{}}
\toprule
 & \multicolumn{2}{c|}{GNEWS} & \multicolumn{2}{c}{20NG} \\ \midrule
 & 50 topics & 100 topics & 50 topics & 100 topics \\\midrule
CTM & 2.1s & 2.2s & 1.2s & 1.2s \\
PLDA & 1.5s & 1.5s & 0.8s & 0.9s \\ \bottomrule
\end{tabular}%
    \caption{Time to complete one epoch on Google News and 20Newsgroups datasets with 25 and 50 topics.}
    \label{tab:my_label}
\end{table}

\end{document}